\documentclass[10pt,twocolumn,letterpaper]{article}

\usepackage[pagenumbers]{cvpr} 

\definecolor{cvprblue}{rgb}{0.21,0.49,0.74}
\usepackage[pagebackref,breaklinks,colorlinks,allcolors=cvprblue]{hyperref}
\usepackage{tikz}
\usepackage{graphicx}
\usepackage{xcolor} 
\usepackage{multirow}

\def\paperID{7085} 
\def\confName{CVPR}
\def\confYear{2026}

\title{PIFF: A Physics-Informed Generative Flow Model for Real-Time Flood Depth Mapping}

\author{ChunLiang Wu\\
Brightest Technology Inc.\\
Hsinchu, Taiwan\\
{\tt\small josh.wu@brightest-tech.com}
\and
Tsunhua Yang\\
National Yang Ming Chiao Tung University\\
Hsinchu, Taiwan\\
{\tt\small tshyang@nycu.edu.tw}
\and 
Hungying Chen \\
National Yang Ming Chiao Tung University\\
Hsinchu, Taiwan\\
{\tt\small hungying.en13@nycu.edu.tw}
}

\begin{document}
\maketitle
\begin{abstract}
Flood mapping is crucial for assessing and mitigating flood impacts, yet traditional methods like numerical modeling and aerial photography face limitations in efficiency and reliability. To address these challenges, we propose PIFF, a physics-informed, flow-based generative neural network for near real-time flood depth estimation. Built on an image-to-image generative framework, it efficiently maps Digital Elevation Models (DEM) to flood depth predictions. The model is conditioned on a simplified inundation model (SPM) that embeds hydrodynamic priors into the training process. Additionally, a transformer-based rainfall encoder captures temporal dependencies in precipitation. Integrating physics-informed constraints with data-driven learning, PIFF captures the causal relationships between rainfall, topography, SPM, and flooding, replacing costly simulations with accurate, real-time flood maps. Using a 26 km² study area in Tainan, Taiwan, with 182 rainfall scenarios ranging from 24 mm to 720 mm over 24 hours, our results demonstrate that PIFF offers an effective, data-driven alternative for flood prediction and response. Code is available at \href{https://github.com/JoshWuuu/PIFF}{https://github.com/JoshWuuu/PIFF}.
\end{abstract}    
\section{Introduction}
\label{sec:intro}

Floods are among the most devastating natural disasters, causing widespread destruction and economic loss \cite{HallFinancialLoss,Doshi2018}. Accurate flood mapping is essential for disaster preparedness, response, and mitigation \cite{AsDCNN}. However, traditional methods such as numerical modeling and aerial photography face significant challenges in terms of computational cost, data availability, and real-time applicability. Numerical models require extensive preprocessing and expert intervention and are usually too slow for most urban-size applications\cite{Bradbrook01092004,chen2007,lhomme2008}. Aerial imagery is usually affected by atmospheric disturbances, such as clouds and fog, which reduce accuracy. Obtaining cloud-free optical satellite images during flooding events is almost impossible\cite{uddin2019operational, li2022satellite}. Research on alternative methods for rapid or near real-time flood mapping has become significant, particularly in data-scarce conditions, to enhance disaster prevention and mitigation efforts \cite{lin2023rapid}.

Recent advancements in artificial intelligence (AI) have opened new possibilities for flood mapping, offering data-driven solutions that can overcome these limitations. \cite{Bent2022,Jones2023} Machine learning techniques can analyze aerial imagery to create accurate maps that inform evacuation planning and disaster relief efforts \cite{AsDCNN,FengRF}. Deep learning models have been employed to predict flood inundation rapidly, significantly reducing computation time compared to traditional hydrodynamic models \cite{PengPSNET,LIselfCNN,NemniCNNSAR,AsDCNN,Hash2021,Marc2019}. Moreover, AI-driven tools have been developed to generate realistic satellite images depicting potential future flooding scenarios, aiding in proactive planning and risk assessment \cite{yokoya2022,guo2021data,Lütjens2024,seo2023improved}. Among these AI approaches, diffusion and flow models \cite{Ho2020DDPM, Song2020DDIM, lipman2022flow, tong2023improving} have emerged as the powerful methods for generating high-quality spatial predictions. By learning probabilistic distributions, they enable realistic and reliable flood simulations without the need for extensive numerical computations. 

To leverage these advancements, we propose PIFF, a novel physics-based generative flow model for near real-time flood mapping. PIFF learns to predict flood depth from digital elevation maps (DEM), 24-hour rainfall time-series data and simplified inundation model (SPM). It adopts a flow matching training objective to estimate the vector field governing the data generation process. For sampling, it employs an ordinary differential equations (ODE) based strategy for deterministic flooding mapping. We validate PIFF using a 26 km² study area in Tainan, Taiwan, training on 182 historical rainfall events and evaluating on 20 rainfall scenarios across different rainfall types, including uniform, non-uniform, and real events. Our results demonstrate that PIFF effectively generates high-quality flood maps, presenting a scalable solution for AI-driven flood prediction and response.

The main contributions of this paper are as follows:
\begin{itemize}
\item We propose PIFF, the first flow matching model for flood depth estimation, showcasing the potential of generative AI for hydrological applications.
\item Our method incorporates a physics-informed approach by conditioning the generative model on a SPM, embedding hydrodynamic knowledge prior to improve prediction accuracy and reliability.
\item We introduce a novel method for encoding time-series rainfall data into the diffusion pipeline, leveraging transformer embeddings and cross-attention mechanisms to capture temporal dependencies for flood prediction.
\end{itemize}
\section{Related Work}
\label{sec:related_work}

\subsection{Traditional Flood Mapping Techniques}
Two-dimensional (2D) hydrodynamic models employing shallow water equations (SWEs) as the governing equations are widely used to generate time-varied flood maps with detailed flood extent and depths \cite{ming2020real, chiang2024efficient}. Although these models yield accurate results, developing complex models is very time-consuming. Additionally, these models require trained personnel for development. Furthermore, computational resources and numerical stability for fully dynamic models present ongoing challenges, particularly for supporting real-time operations during emergency responses. Alternatives have been developed to enhance calculation efficiency. For instance, utilizing a Graphics Processing Unit (GPU) can significantly improve real-time, high-resolution flood modeling \cite{ming2020real}. Alternatively, one could simplify the governing equations to sacrifice some prediction accuracy \cite{wijaya2023rapid}. The rise of AI has garnered considerable attention for boosting 2D calculation efficiency \cite{chiang2024efficient}. None of the above has been chosen as the best option for real-time flood mapping. Using a GPU can save computational time, but the preprocessing time for building model complexity is still significant. Additionally, it requires well-trained personnel. Simplifying governing equations speeds up the calculation process but the prediction accuracy cannot always meet the requirement of emergency operation. Finally, AI is a powerful tool for flood modeling, but current applications still focus on the causal relationships between input factors and output results\cite{chiang2024efficient, kumar2025machine}. Accurate predictions and forecasts from AI-based flood models require reliable, high-resolution spatial-temporal data inputs. However, such datasets may not always available \cite{kumar2025machine}.

\subsection{AI-Based Flood Mapping}
Advancements in AI have transformed flood mapping by enabling data-driven models to learn complex flood patterns and enhance prediction accuracy. Research in this area can be categorized into two key approaches: flood segmentation and flood depth estimation. 

\subsubsection{AI-Based Flood Segmentation}
The first approach, flood segmentation, focuses on identifying and delineating flooded areas within an image, typically formulated as an image segmentation task. Deep learning models, particularly convolutional neural networks (CNNs) with encoder-decoder architectures, are widely used to extract spatial features and generate pixel-wise flood masks. Wieland \cite{Marc2019} used an encoder-decoder CNN to segment floodwater, permanent water bodies, and other contextual elements like snow, land, and clouds from satellite imagery. PSNET \cite{PengPSNET} utilizes CNNs to analyze patches of pre- and post-flood images from the same area, determining whether a region is flooded. Nemni \cite{NemniCNNSAR} proposed a CNN-based approach for flood segmentation in Synthetic Aperture Radar (SAR) imagery, achieving an 80\% reduction in flood map generation time. Hashemi-Beni \cite{Hash2021} introduced a hybrid model combining CNNs with a region-growing method to segment both visible and submerged flooded areas, including those obscured by vegetation. FloodGAN \cite{floodgan2021} is based on generative adversarial network (GAN) to generate flooded map conditioned by rainfall map. The results demonstrate that the proposed floodGAN model is up to 106 times faster than the hydrodynamic model and promising in terms of accuracy and generalizability. To overcome Electro-Optical (EO) satellite limitations like cloud cover and poor nighttime visibility, Seo et al. \cite{seo2023improved} proposed a diffusion-based framework to convert Synthetic SAR images into EO imagery for improved flood segmentation and interpretability.

While these methods effectively segment flooded regions, they primarily focus on flood segmentation rather than generating flood depth maps, which are crucial for accurate risk assessment and response planning.

\subsubsection{AI-Based Flood Depth Estimation}
The second approach, flood depth estimation, aims to generate flood maps that not only delineate flood extent but also predict flood depth at each location. Unlike segmentation-based methods, these models provide a more detailed representation of flooding severity. Yokoya et al. \cite{yokoya2022} implemented a U-Net with attention blocks to predict maximum water levels from a combination of pre- and post-disaster satellite images and digital elevation maps (DEM). Similarly, Guo et al. \cite{guo2021data} formulated flood depth estimation as an image-to-image translation problem using CNNs, where rainfall hyetographs and terrain data serve as inputs to generate maximum water depth rasters. U-FLOOD \cite{LOWE2021126898} follows an encoder-decoder CNN structure. Rainfall series are transformed into nine statistical features, which are fed into a fully connected layer and fused with spatial convolutions at the network’s bottleneck to enhance flood depth prediction. U-Net$_{River}$ \cite{HOSSEINY2021} takes ground elevation and flooding discharge as input, demonstrating a 29\% improvement in maximum flood depth prediction within rivers compared to traditional hydraulic simulations. Beyond CNN-based methods, Lütjens et al. \cite{Lütjens2024} leveraged GAN to synthesize post-flood satellite imagery from pre-flood images and physics-based flood maps, aiding in future flood event simulation. 

Despite these advancements, most existing approaches rely on encoder-decoder architectures and GANs for flood depth generation, with limited research exploring more advanced generative models for this task. Moreover, few studies incorporate temporal rainfall data into the generative process. Our work builds upon this research by conditioning the flood depth generation process on 24-hour rainfall time series.
\section{Method}
\label{sec:method}
\usetikzlibrary{positioning}
\usetikzlibrary{shapes.geometric, arrows}
\tikzstyle{arrow} = [thick,->,>=stealth]
\tikzstyle{straight} = [thick,-,>=stealth]
\tikzstyle{dashedarrow} = [thick,dashed,->,>=stealth]
\tikzstyle{dottedline} = [thick, dotted, dash pattern=on 3pt off 2pt]
\tikzstyle{circlearrow} = [circle, draw, minimum size=0.4cm]
\begin{figure*}[ht]
\centering
\vskip 0.2in
\begin{tikzpicture}[node distance=0.7cm]
\draw[thick, red!30, fill=red!10, rounded corners=5pt] (16.5, 6.2) rectangle (1.5, 2.7);




\node at (10.2, 5.6) {\fontsize{8}{8} \textbf{Forward Diffusion}};
\node[draw, black, thick ,inner sep=0] at (2.9,5.3) {\includegraphics[width=1.3cm]{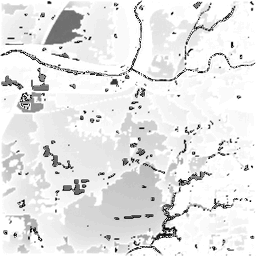}};
\node at (1.8, 5.3) {\fontsize{12}{12} \textbf{$x_0$}};
\draw[arrow] (3.55,5.3) -- (4.3,5.3);

\node[draw, fill=orange!10, thick, rounded corners, font=\large,inner sep=7pt] at (4.8,5.3) {$x_{t1}$};
\node[draw, fill=orange!10, thick, rounded corners, font=\large,inner sep=7pt] at (6.2,5.3) {$x_{t2}$};
\node[draw, fill=orange!10, thick, rounded corners, font=\large,inner sep=7pt] at (7.6,5.3) {$x_{t3}$};
\node[draw, fill=orange!10, thick, rounded corners, font=\large,inner sep=7pt] at (13.2,5.3) {$x_{tx}$};
\draw[arrow] (5.3, 5.3) -- (5.7, 5.3);
\draw[arrow] (6.7,5.3) -- (7.1,5.3);
\draw[dottedline] (8.1,5.3) -- (11,5.3);
\draw[arrow] (11,5.3) -- (12.7,5.3);

\node[draw, black, thick,inner sep=0] at (15.1 ,5.3) {\includegraphics[width=1.3cm]{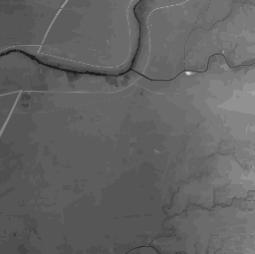}};
\node at (16.1, 5.3) {\fontsize{12}{12} \textbf{$x_1$}};
\draw[arrow] (13.7,5.3) -- (14.45,5.3);

\node at (5.2, 4.2) {\fontsize{8}{8} \textbf{Backward Generation}};
\node[draw, black, thick,inner sep=0] at (15.1 ,3.5) {\includegraphics[width=1.3cm]{sec/figs/dem_gray.png}};
\node at (16.1, 3.5) {\fontsize{12}{12} \textbf{$\hat{x}_1$}};
\draw[arrow] (14.45,3.5) -- (13.7,3.5);

\node[draw, fill=orange!10, thick, rounded corners, font=\large,inner sep=7pt] at (13.2,3.5) {$\hat{x}_{t1}$};
\node[draw, fill=orange!10, thick, rounded corners, font=\large,inner sep=7pt] at (4.8,3.5) {$\hat{x}_{tx}$};

\node[draw, black, thick ,inner sep=0] at (2.9,3.5) {\includegraphics[width=1.3cm]{sec/figs/piff/contrast_133_20_gen.png}};
\node at (1.8, 3.5) {\fontsize{12}{12} \textbf{$\hat{x}_0$}};
\draw[arrow] (4.3,3.5) -- (3.55,3.5);

\draw[thick, red!30, fill=red!20, rounded corners=5pt] (11.7, 4.6) rectangle (6.7, 2.6);
\node at (9.3, 4.4) {\fontsize{8}{8} \textbf{Denoised Network $\epsilon_\theta$}};
\draw[thick, fill=orange!20] (7.1, 2.7) -- (9.0,3.0) -- (9.0, 4.0) -- (7.1,4.2) -- cycle;
\draw[thick, fill=orange!20] (9.4, 3.0) -- (11.3,2.7) -- (11.3, 4.2) -- (9.4,4.0) -- cycle;
\node[draw, fill=brown!10, thick, rounded corners, font=\small,inner sep=3pt, rotate=90] at (10.3,3.5){$QKV$};
\node[draw, fill=brown!10, thick, rounded corners, font=\small,inner sep=3pt, rotate=90] at (8.1,3.5) {$QKV$};
\draw[straight] (9.0, 3.5) -- (9.4, 3.5);

\draw[straight] (12.7, 3.5) -- (11.7, 3.5);
\draw[arrow] (6.7,3.5) -- (5.3,3.5);

\draw[thick, blue!30, fill=blue!10, rounded corners=5pt] (7.8, 2.5) rectangle (4.5, 1);
\node at (5.3,2.3) {\fontsize{8}{8} \textbf{Conditioning}};
\node[draw, fill=purple!15, thick, rounded corners, font=\small,inner sep=7pt] at (5.5,1.75) {Rainfall $r$};
\draw[thick, fill=yellow!20] (7.0, 1.15) -- (7.7,1.4) -- (7.7, 2.1) -- (7.0,2.35) -- cycle;
\node at (7.35,1.75) [align=center] {$H$};
\draw[arrow] (6.4,1.75) -- (7.0,1.75);
\draw [arrow] (7.7, 1.75) -- ++(0.4,0) -- ++(0,1.25);
\draw [arrow] (7.7, 1.75) -- ++(2.6,0) -- ++(0,1.25);

\draw[thick, blue!30, fill=blue!10, rounded corners=5pt] (12.5, 2.5) rectangle (14.4, 1);
\node at (13.3,2.3) {\fontsize{8}{8} \textbf{Conditioning}};
\node[draw, fill=purple!15, thick, rounded corners, font=\small,inner sep=7pt] at (13.3,1.75) {SPM $s$};
\draw [arrow] (12.5, 1.75) -- ++(-0.4,0) -- ++(0,1.25) -- ++(-0.4,0);

\end{tikzpicture}
\caption{Overview of PIFF's forward and backward processes: the forward diffusion (top) progressively diffuse DEM image $x_0$, whereas the backward process (bottom) iteratively denoises $\hat{x}_1$ to reconstruct $\hat{x}_0$, which is then decoded to obtain the output $\hat{x}_0$}
\label{fig:PIFF flow}
\vskip -0.2in
\end{figure*}

PIFF is a novel flow-based model designed for flood depth estimation, leveraging the optimal transport conditional flow matching (OT-CFM) \cite{tong2023improving} framework for efficient mapping between DEM and flood depth predictions. Unlike standard diffusion/flow models that generate outputs from Gaussian noise, OT-CFM directly learns the vector field between source and target distributions, making it particularly effective for structured prediction tasks such as flood modeling. 
To improve training efficiency, our method incorporates a physics-informed approach by conditioning the generative model on SPM, which embeds domain-specific hydrodynamic knowledge into the learning process. Additionally, we introduce a time-series rainfall encoding mechanism, utilizing transformer embeddings and cross-attention mechanisms \cite{vaswani2017attention}, to capture temporal dependencies essential for accurate flood forecasting.

\subsection{PIFF Training}

PIFF employs the flow matching framework to learn a continuous transformation between the digital elevation model (DEM) and the corresponding flood depth map. Specifically, the model is trained to interpolate between the two distributions by modeling the conditional trajectory \(x_t\) at each time step \(t \in [0,1]\), as illustrated in the top half of \cref{fig:PIFF flow}.

The conditional distribution of \(x_t\) given the flood map \(x_0\) and DEM \(x_1\) is defined as:
\begin{equation}
\label{eq:piff_cond}
p_t(x \mid x_1, x_0) = \mathcal{N}(x \mid (1-t)x_0 + t x_1, \sigma^2 I)
\end{equation}

At each training step, PIFF randomly samples \(t \in [0,1]\) and generates an intermediate state \(x_t\) by linearly interpolating between the flood map (\(x_0\)) and the DEM (\(x_1\)), with added Gaussian noise.

The corresponding velocity (vector field) at time \(t\) is obtained by taking the partial derivative of the intermediate state \(x_t\) with respect to \(t\):
\begin{equation}
\label{eq:piff_velocity}
u_t(x \mid x_1, x_0) = x_1 - x_0
\end{equation}

The model, parameterized by \(\theta\), predicts the velocity field \(u_\theta(x_t, x_1, t, r, s)\) conditioned on rainfall \(r\), DEM \(x_1\), SPM \(s\), timestep \(t\). The Conditional Flow Matching (CFM) loss is then defined as the mean squared error between the predicted and true velocity fields:
\begin{equation}
\label{eq:cfm_loss}
\mathcal{L}_{\text{CFM}} = \mathbb{E}_{x_0, x_1, r, s, t}\left[\left\|u_\theta(x_t, x_1, t, r, s) - (x_1 - x_0)\right\|^2\right]
\end{equation}

This loss encourages the network to learn the optimal flow between DEM and flood map, enabling robust and efficient generation of flood depth predictions from input DEMs. 



\subsection{PIFF Sampling}

In the backward process, PIFF gradually transforms the DEM into the corresponding flood depth map by reversing the learned generative flow, as illustrated in the bottom half of \cref{fig:PIFF flow}. For inference, we formulate the generation as solving an Ordinary Differential Equation (ODE) that traces the learned vector field from the initial DEM (\(x_1\)) towards the target flood depth prediction.

Specifically, given the trained velocity field \(u_\theta\) conditioned on rainfall, DEM, and SPM, the intermediate state \(\hat{x}_t\) evolves according to:
\begin{equation}
\frac{d \hat{x}_t}{d t} = u_\theta(\hat{x}_t, t, r, x_1, s)
\end{equation}

To numerically integrate the ODE and recover the flood depth map, we employ the standard Euler method. This ODE solver updates \(\hat{x}_t\) iteratively over \(n\) steps with a specified step size \(h\), offering a balance between computational efficiency and accuracy.

\subsection{Physics-Informed Prior via SPM}
The Simplified Inundation Model (SPM) \cite{yang2015spm} is a DEM-based inundation model that incorporates fundamental physical principles to simulate flood depths across a spatial domain. It assumes that rainfall uniformly accumulates as an incremental flood depth ($\Delta h_1$) at each grid cell within the study area.$Z_1$ represent the elevation at a given grid cell. When the accumulated water depth at $Z_1$ exceeds the elevation of an adjacent grid cell $Z_2$, i.e.,
\begin{equation}
Z_1 + \Delta h_1 > Z_2
\end{equation}
floodwater is permitted to flow into $Z_2$. The incremental flood depth transferred to $Z_2$, denoted as $\Delta h_2$, is calculated as:
\begin{equation}
\Delta h_2 = (Z_1 + \Delta h_1 - Z_2) \times \Phi
\end{equation}
Here, $\Phi$ is a constant representing the proportion of flood volume transferred from $Z_1$ to $Z_2$. The model adopts a D4 flow scheme, in which water is distributed equally to the four neighboring grid cells, thus $\Phi = 0.25$.

The process is iteratively applied to all relevant grid cells. The simulation continues until the flood depth $\Delta h_i$ at each grid cell ($i = 1, 2, 3, \dots, n$) meets or exceeds the prescribed reference rainfall depth $r$, serving as a mass balance constraint:
\begin{equation}
\Delta h_i \geq \text{Reference rainfall depth } r
\end{equation}
In this study, the reference rainfall depth corresponds to the total accumulated rainfall ($r$ mm) over a 24-hour period. This condition ensures that the simulated flood volume conserves mass by matching the total rainfall input. In summary, the inputs to the SPM model include the DEM and rainfall depth $r$, while the mass conservation condition ensures that the cumulative simulated flood volume equals the rainfall uniformly distributed across the study area.

In our framework, SPM acts as a physics-informed prior, providing essential hydrodynamic context while remaining lightweight. For each timestep during PIFF’s training and sampling, the corresponding SPM output—based on the 24-hour rainfall summation up to that point—is concatenated with the intermediate model state \(x_t\). This design ensures that the model conditions its predictions on both current data-driven features and physically plausible flood patterns derived from recent rainfall dynamics.

\subsection{Time-Series Rainfall Encoding with Cross-Attention Mechanism}

PIFF incorporates the past 24 hours time-series rainfall data as a conditioning signal, encoding temporal dependencies to improve flood depth predictions. The rainfall time series $\{R_t\}_{t=1}^{24}$ is first encoded using a transformer encoder, which is chosen over other encoding methods due to its cross-attention and timestep embedding mechanisms. These features enhance the encoding by effectively capturing the mutual relationships between rainfall events in the time series:

\begin{equation}
    H = \text{TransformerEncoder}(R)
\end{equation}
                           
We then integrate its embeddings, $H$, into the diffusion process via cross-attention, injecting temporal information into intermediate representations $z$ of the denoised model $\epsilon_\theta$.
\section{Experiments}
\label{sec:experiments}
\subsection{Experimental Setup}
\begin{figure}[t]
    \centering
    \includegraphics[width=0.6\linewidth]{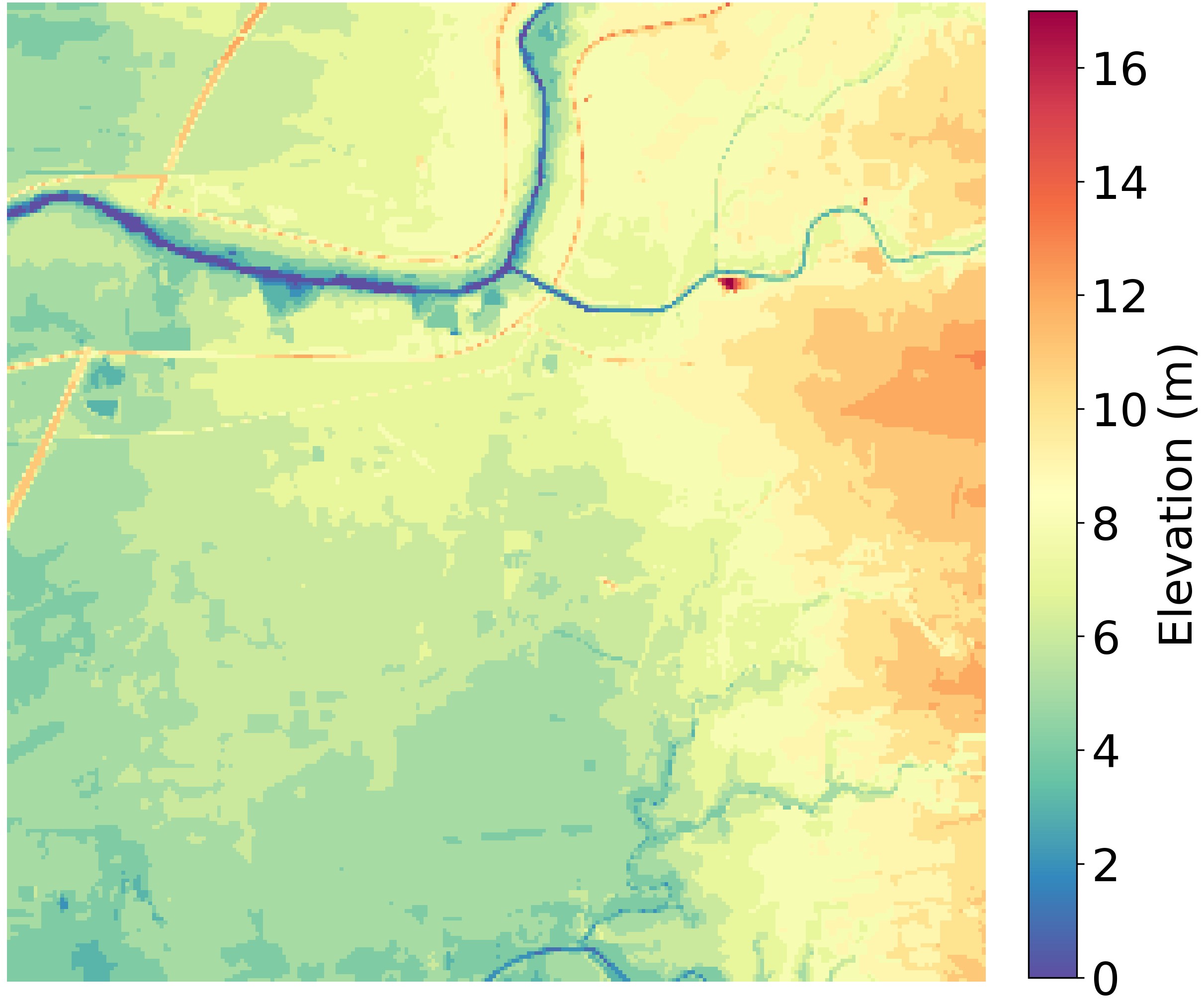}
    \caption{DEM image from Tainan, Taiwan. The study area includes a channel in the northern part, and the overall ground slope decreases gently from east to west. The average terrain slope is approximately 0.014, with a total elevation difference of about 10 m across the area.}
    \label{fig::dem}
\end{figure}

\paragraph{Datasets} Since it is nearly impossible to obtain detailed aerial observations of flood extent in real-world scenarios, the flood data used as ground truth in this study is generated using the TUFLOW simulation model \cite{tuflow}, based on a 26 km² digital elevation model (DEM) from Tainan, Taiwan. DEM is obtained from coordinate EPSG:3826 TWD97 in \href{https://data.gov.tw/dataset/35430}{Taiwan Government's Open Data Platform}. \cref{fig::dem} shows the DEM used for simulations. In the example simulated images (top-row image in \cref{fig::chrono_figs}), the flood extent corresponds to areas with pixel values below 255, while darker pixels indicate deeper flood depths. The dataset comprises 256×256 pixel images, with each pixel representing a 20-meter spatial resolution.

\begin{figure}[t]
\centering
\hspace*{-0.4cm}
    \includegraphics[width=8.7cm]{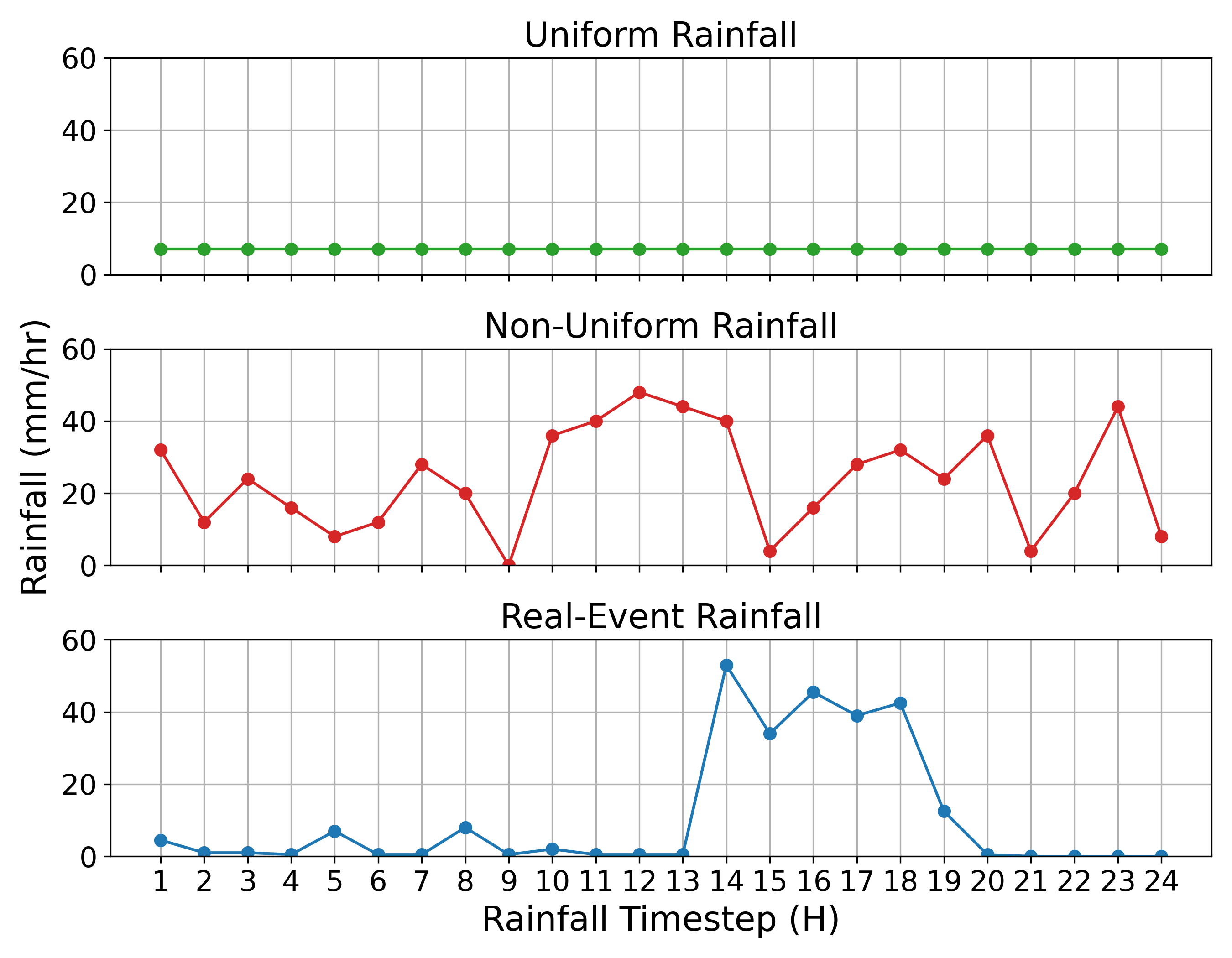}
    \caption{Comparison of uniform, non-uniform, and real-event rainfall patterns over a 24-hour duration}
    \label{fig::rainfall}
\end{figure}

The training rainfall dataset includes a total of 182 distinct scenarios, categorized into uniform, non-uniform, and real event rainfall conditions. 
\begin{itemize}
    \item Uniform rainfall: this category consists of 10 scenarios where the rainfall intensity is constant over the entire 24-hour period. Each scenario has a different constant intensity.
    \item Non-uniform rainfall: these 45 scenarios are generated from a Gaussian distribution over 24 hours, and the hourly rainfall values are then randomly shuffled.
    \item Real-event rainfall: this category includes 127 historical rainfall events in Taiwan. These events feature distinct fluctuations that reflect natural storm patterns. The data is sourced from station C0X060 in \href{https://codis.cwa.gov.tw/StationData}{Climate Observation Data Inquire Service (CODiS)}
\end{itemize}
For testing, the dataset comprises 20 rainfall scenarios, consisting of 2 uniform rainfall events, 3 non-uniform events, and 15 real-event cases. Each rainfall scenario spans 25 hours, generating 4,050 images for training and 500 images for testing. \cref{fig::rainfall} shows three distinct rainfall patterns.

\paragraph{Hyperparameters} a noise level of $\sigma = 0.1$, batch size of 128, and an initial learning rate of $5 \times 10^{-4}$. The model is optimized for 10{,}000 iterations using a linear learning rate decay schedule, where the rate is reduced by a factor of 0.99 every 1{,}000 steps. The model is trained on RTX4090.

\paragraph{Baselines}We evaluate PIFF against two baseline categories. The first is a physics-based simulator (SPM \cite{yang2015spm}). The second includes AI-driven generative models: a U-Net \cite{ronneberger2015unet}, a Generative Adversarial Network (GAN) \cite{isola2018gan}, and a flow matching (FM) model \cite{lipman2022flow}. In all AI-based models, rainfall data is encoded via a transformer encoder. In the FM baseline, the target distribution $x_1$ is set as Gaussian noise, while in PIFF, the target distribution corresponds to the DEM. 

\paragraph{Evaluation} We evaluate our model using both qualitative and quantitative experiments. We first qualitatively assess the chronological progression of generated flood images from $t=4$ to $t=24$. This inspection uses the real-world rainfall time series from Typhoon Gaemi (July 26, 2025, Tainan, Taiwan), as depicted in \cref{fig::rainfall}. We next perform a comprehensive evaluation on uniform, non-uniform, and real-event rainfall scenarios. This assessment is split into two parts. Firstly, we use computer vision metrics (L1, L-infinity, FID \cite{heusel2017gans}) and flood-specific metrics, mean average error(MAE) and maximum difference(MD). The flood depth MD is the L1 distance between the ground truth's maximum flood depth and the predicted value at that same location. Secondly, to verify flood event capture rate, we use a confusion matrix based on a 30 cm flood identification threshold \cite{VisualizingMultipleMeasuresofForecastQuality}, a standard empirical value for flood events in Taiwan. The components are defined as:
\begin{itemize}
    \item Hit (True Positive, $a$): Both observed and predicted water depths are $\ge$ 30 cm.
    \item Miss (False Negative, $b$): Observed water depth is $\ge$ 30 cm, but predicted water depth is $<$ 30 cm.
    \item False alarm (False Positive, $c$): Predicted water depth is $\ge$ 30 cm, but observed water depth is $<$ 30 cm.
    \item Correct negative (True Negative, $d$): Both observed and predicted water depths are $<$ 30 cm.
\end{itemize}
Based on these definitions, we calculate common scores like the probability of detection (POD), false-alarm rate (FAR), accuracy, bias, and critical success index (CSI) as follows:

\begin{align*}
POD (Recall) &= \frac{a}{a+b}, \quad
FAR (1-Precision) = \frac{c}{a+c}, \\
Bias &= \frac{a+c}{a+b}, \quad
CSI = \frac{a}{a+b+c}
\end{align*}

\subsection{Experimental Results}

\begin{figure}[t]
    \begin{tikzpicture}

        \foreach \i [count=\j from 0] in {12,16,20} {
            \node[draw=gray, thick, anchor=south west, inner sep=0] (GT2\j) at (\j*2.65, 5*2.65) 
                {\includegraphics[width=2.6cm]{sec/figs/gt/contrast_133_\i_gt.png}};
            \node[draw=gray, thick, anchor=south west, inner sep=0] (piff\j) at (\j*2.65, 4*2.65) 
                {\includegraphics[width=2.6cm]{sec/figs/piff/contrast_133_\i_gen.png}};
            \node[draw=gray, thick, anchor=south west, inner sep=0] (fm\j) at (\j*2.65, 3*2.65) 
                {\includegraphics[width=2.6cm]{sec/figs/fm/133_\i.png}};
            \node[draw=gray, thick, anchor=south west, inner sep=0] (gan\j) at (\j*2.65,2*2.65) 
                {\includegraphics[width=2.6cm]{sec/figs/gan/133_\i.png}};
            \node[draw=gray, thick, anchor=south west, inner sep=0] (unet\j) at (\j*2.65, 1*2.65) 
                {\includegraphics[width=2.6cm]{sec/figs/unet/133_\i.png}};
            \node[draw=gray, thick, anchor=south west, inner sep=0] (spm\j) at (\j*2.65, 0) 
                {\includegraphics[width=2.6cm]{sec/figs/spm/133_\i.png}};
            \node[above] at (GT2\j.north) {\textbf{t=\i}};
        };

        \node[left, rotate=90] at (-0.2, 15.6) {\footnotesize \textbf{Ground Truth}};
        \node[left, rotate=90] at (-0.2, 12.4) {\footnotesize \textbf{PIFF}};
         \node[left, rotate=90] at (-0.2, 9.6) {\footnotesize \textbf{FM}};
        \node[left, rotate=90] at (-0.2, 7.0) {\footnotesize \textbf{GAN}};
         \node[left, rotate=90] at (-0.2, 4.5) {\footnotesize \textbf{U-Net}};
                \node[left, rotate=90] at (-0.2, 1.8) {\footnotesize \textbf{SPM}};
        
        \pgfmathsetmacro{\colorbarXPos}{-0.2} 
        \node[anchor=south west, inner sep=0] at (\colorbarXPos, -1.0)
        {\includegraphics[height=1cm, width=8.4cm]{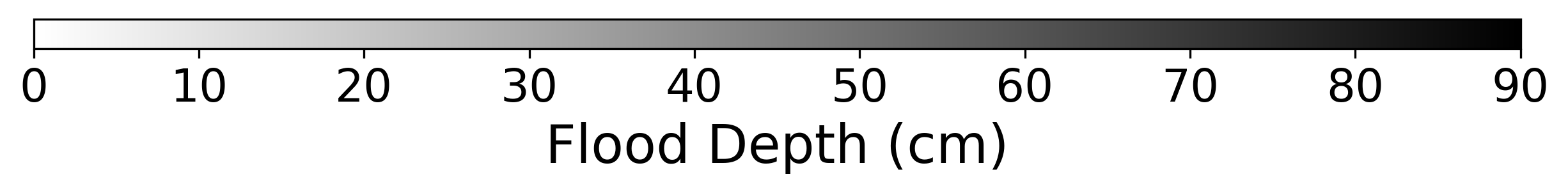}};
    \end{tikzpicture}
    \caption{Comparison of ground truth and PIFF's generated flooded images at different timesteps ($t$) from $t=12$ to $t=20$ in \cref{fig::rainfall}. Note: Image contrast is enhanced using min-max normalization with $min = 200$ and $max = 255$.}
    \label{fig::chrono_figs}
\end{figure}

\paragraph{PIFF accurately captures flood dynamics and depth} \cref{fig::chrono_figs} shows the chronological flood progression from t=12 to t=20, driven by the real-event rainfall in \cref{fig::rainfall}. Following a period of low rainfall with minimal flooding (t=12), a sharp intensity increase after hour 13 causes a progressive inundation, visible at t=16. Throughout this temporal evolution, PIFF's predictions strongly align with the ground truth. A detailed comparison at the 16th hour further highlights PIFF's superiority. While both PIFF and FM capture the overall flood extent well, FM consistently underestimates the flood depth (indicated by lighter pixels). In contrast, PIFF accurately models both the flood's extent and depth. 

\paragraph{PIFF achieves superior performance in all metrics} The quantitative results in \cref{tab:quantitative results l1 linf} show that AI-based methods consistently outperform the physics-based method. Among the AI models, a clear trend emerges: the diffusion-like frameworks (FM and PIFF) are superior to U-Net and GAN. Building on this, PIFF which integrates physics-informed constraints sets the highest benchmark. It leads in all image metrics (L1, L-inf, FID) and flood metrics (MAE, MD) across all three rainfall scenarios. Its performance is particularly strong in the challenging real-event case (L1: 0.2, FID: 2.3), underscoring its promise for real-world application. Furthermore, PIFF's drastically reduced MD of just 7.2 in the real event highlights a key benefit: the physics-informed constraints successfully guide the model to minimize large errors in flood depth prediction.

\begin{table*}[t]
\caption{Image metric(L1, L-inf, FID) and flood metric(MAE and MD) for algorithm performance on uniform, non-uniform, and real-event rainfalls}
\label{tab:quantitative results l1 linf}
\begin{center}
\begin{small}
\begin{sc}
\begin{tabular}{ c l cc ccc}
\toprule
 &  & \multicolumn{3}{c}{Image Metric} & \multicolumn{2}{c}{Flood Metric} \\ \cmidrule(lr){3-5} \cmidrule(lr){6-7}
Rainfall Type & Method & L1(↓) & L-inf(↓) & FID(↓)  & MAE(↓) & MD(↓) \\ 
\midrule
\multirow{5}{*}{\textit{Uniform}} &
SPM \cite{yang2015spm} & 2.5 & 175.0 & 74.4 & 3.8 & 76.8 \\
&U-Net \cite{ronneberger2015unet} & 1.6 & 71.1 & 73.7 & 2.5 & 60.2 \\
&GAN \cite{isola2018gan} & 1.3 & 66.9 & 59.0 & 2.2 & 56.0 \\
&FM \cite{lipman2022flow} & 0.9 & 47.3 & 19.9 & 1.4 & 16.9 \\
&PIFF (Ours) & \textbf{0.3} & \textbf{23.4} & \textbf{8.6} & \textbf{0.4} & \textbf{13.8} \\
\midrule
\multirow{5}{*}{\textit{Non-Uniform}} &
SPM \cite{yang2015spm} & 7.4 & 205 & 110.5 & 11.6 & 59.8 \\
&U-Net \cite{ronneberger2015unet} & 2.9 & 110.3 & 70.6 & 4.5 & 84.8 \\
&GAN \cite{isola2018gan} & 2.7 & 97.0 & 63.0 & 4.2 & 71.2  \\
&FM \cite{lipman2022flow} & 1.2 & 41.0 & 18.5 & 1.8 & 32.4 \\
&PIFF (Ours) & \textbf{0.8} & \textbf{36.1} & \textbf{9.3} & \textbf{1.3} & \textbf{23.2} \\
\midrule
\multirow{5}{*}{\textit{Real Event}} &
SPM \cite{yang2015spm} & 2.2 & 151.8 & 47.0 & 3.4 & 53.6  \\
&U-Net \cite{ronneberger2015unet} & 1.4 & 58.8 & 65.8 & 2.0 & 51.3  \\
&GAN \cite{isola2018gan} & 1.2 & 58.4 & 40.7 & 1.8 & 42.1  \\
&FM \cite{lipman2022flow} & 0.5 & 21.1 & 20.1 & 0.9 & 17.0 \\
&PIFF (Ours) &  \textbf{0.2} & \textbf{14.3} & \textbf{2.3} & \textbf{0.2} & \textbf{7.2} \\
\bottomrule
\end{tabular}
\end{sc}
\end{small}
\end{center}
\end{table*}

\begin{figure*}[t]
    \centering
    \includegraphics[width=1\linewidth]{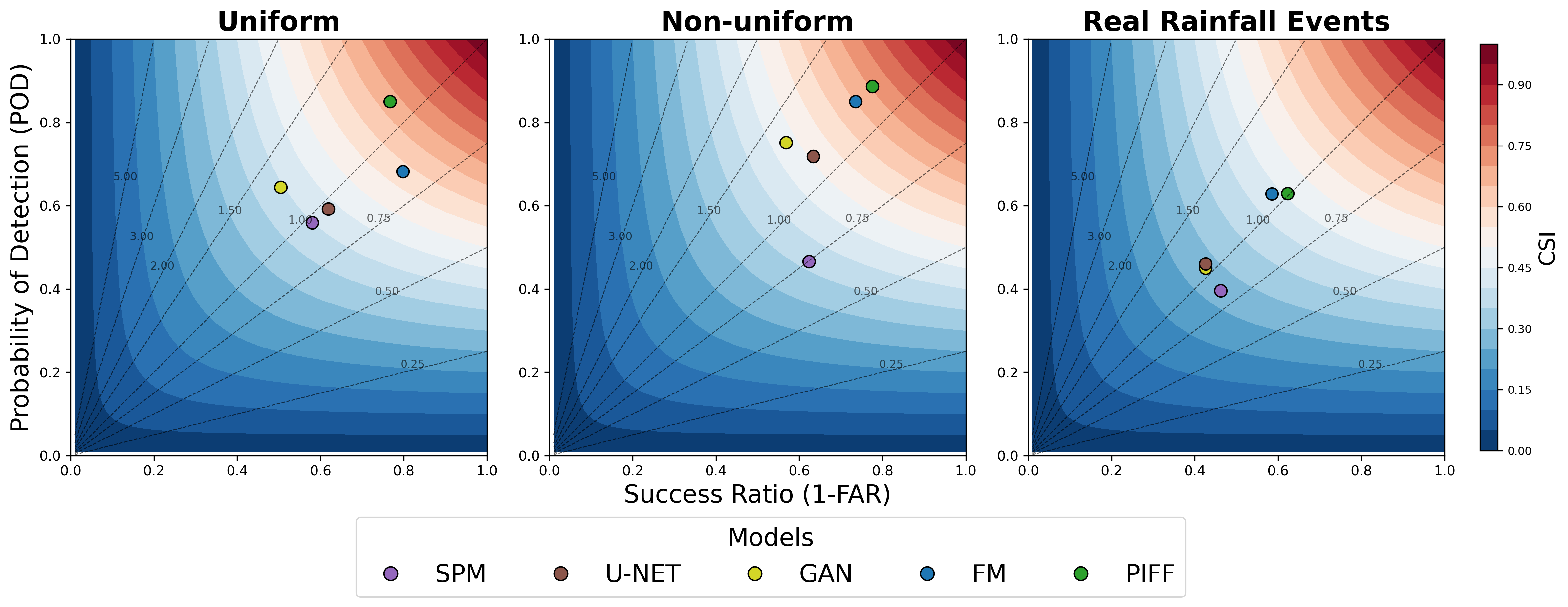}
    \caption{Four-metric (POD, 1-FAR, Bias, CSI) Plot for Algorithm Performance on Uniform, Non-uniform, and Real Event Rainfalls}
    \label{fig::confusion matrix}
\end{figure*}

\paragraph{PIFF achieves superior classification accuracy and precision} \cref{fig::confusion matrix} illustrates the models' categorical performance using a four-metric plot (POD, 1-FAR, Bias, CSI). In this visualization, the optimal performance is represented by the top-right corner (i.e., perfect POD and 1-FAR). PIFF's dot consistently lies closest to this ideal point for all uniform, non-uniform, and real-event rainfall scenarios. This indicates that PIFF is the most accurate model for classifying flooding events (water depth $\geq$ 30cm).

For the critical real-event rainfall scenario, PIFF's high 1-FAR score is particularly significant. This metric is equivalent to precision (1 - false alarm rate), which is a vital measure for real-world flood management. A high-precision model ensures that when a flood is predicted, it is highly likely to actually occur. This reliability is crucial for emergency response, enabling the confident allocation of urgent relief resources and minimizing costly false alarms.

\subsection{Discussion}
\paragraph{SPM prior guidance improves all AI Models} \cref{tab:spm prior on all models} presents a study on the impact of the SPM physics prior, focusing on the real-event rainfall scenario. The results show that incorporating this prior provides a consistent and significant benefit to all tested AI models, from U-Net to our proposed PIFF. This demonstrates that the physics-based guidance is not redundant but provides crucial constraints that improve accuracy and mitigate worst-case errors (L-inf and MD) across all architectures.

\paragraph{A direct DEM-to-flood diffusion path improves performance} \cref{tab:spm prior on all models} allows for a direct comparison between FM with SPM prior and PIFF, as both are flow matching frameworks benefiting from the same SPM prior. The key architectural difference lies in their target distributions: FM diffuses from a standard Gaussian distribution, whereas PIFF diffuses from the DEM image. The results show PIFF is superior in every metric. The most dramatic difference is in FID, where PIFF (2.3) achieves significantly better perceptual quality than FM (19.6). This strongly suggests that setting the diffusion path on a relevant physical map (the DEM) provides a more effective and constrained mapping than diffusing from Gaussian noise.

\definecolor{darkred}{rgb}{0.6, 0.0, 0.0}

\begin{table*}[t]
\caption{Ablation study on the impact of the SPM physics prior on AI model performance for the real-event rainfall scenario.}
\label{tab:spm prior on all models}
\begin{center}
\begin{small}
\begin{sc}
\begin{tabular}{ c c cc ccc}
\toprule
 &  & \multicolumn{3}{c}{Image Metric} & \multicolumn{2}{c}{Flood Metric} \\ \cmidrule(lr){3-5} \cmidrule(lr){6-7}
Method & SPM prior & L1(↓) & L-inf(↓) & FID(↓)  & MAE(↓) & MD(↓) \\ 
\midrule
U-Net & \checkmark &  1.1 & 48.8 & 35.8 & 1.6 & 22.0  \\
& $\times$ & \textcolor{darkred}{+0.3} & \textcolor{darkred}{+10.0} & \textcolor{darkred}{+30.0} & \textcolor{darkred}{+0.4} & \textcolor{darkred}{+29.3} \\
GAN  & \checkmark &  1.1 & 57.1 & 31.7 & 1.7 & 41.7  \\
 &$\times$ & \textcolor{darkred}{+0.1} & \textcolor{darkred}{+1.3} & \textcolor{darkred}{+9.0} & \textcolor{darkred}{+0.1} & \textcolor{darkred}{+0.4} \\
FM & \checkmark &  0.4 & 16.0 & 19.6 & 0.7 & 7.5 \\
 & $\times$ & \textcolor{darkred}{+0.1} & \textcolor{darkred}{+5.1} & \textcolor{darkred}{+0.5} & \textcolor{darkred}{+0.2} & \textcolor{darkred}{+9.5} \\
PIFF & \checkmark & 0.2 & 14.3 & 2.3 & 0.2 & 7.2 \\
 & $\times$ & \textcolor{darkred}{+0.1} & \textcolor{darkred}{+4.0} & \textcolor{darkred}{+0.2} & \textcolor{darkred}{+0.1}& \textcolor{darkred}{+8.4} \\
\bottomrule
\end{tabular}
\end{sc}
\end{small}
\end{center}
\end{table*}

\paragraph{Rainfall transformer encoder}
We also investigated the influence of the rainfall transformer encoder, with the results shown in \cref{tab:rainfall encoder}. The results demonstrate that this component is critical to the model's performance. The version without the encoder performs significantly worse across all metrics. For example, L1 error increases from 0.2 to 1.0, and the FID score degrades from 2.3 to 40. This improvement is especially pronounced in the flood metrics, where the Maximum Deviation (MD) drops from 32.9 to 7.2, highlighting the encoder's essential role in accurately processing the rainfall time series.

\begin{table*}[t]

  \caption{Ablation study on the effectiveness of the rainfall transformer encoder for the real-event rainfall scenario.}
  \centering
  \begin{tabular}{@{}cccccccc@{}}
    \toprule
     rainfall transformer encoder & L1 & L-inf & FID & flood depth MAE & flood depth MD \\
    \midrule
     $\times$ & 1.0 & 33.8 & 40 & 1.5 & 32.9\\
     $\checkmark$ & 0.2 & 14.3 & 2.3 & 0.2 & 7.2 \\
    \bottomrule
  \end{tabular}
  \label{tab:rainfall encoder}
\end{table*}

\paragraph{Analysis of ODE Solver Performance}
\cref{tab:ode} presents a comparison between the Euler, Heun, and RK4 ODE solvers. The L1 scores are identical for all methods across all rainfall types, demonstrating no difference in average error. We observed very minor variations in MD metric. For the real-event scenario, RK4 performed slightly better than Heun and Euler. However, considering that higher-order solvers like RK4 incur a substantial computational cost for this negligible gain, we concluded that the computationally efficient Euler method provides the best practical trade-off.
\begin{table*}[t]
\caption{Performance comparison of different ODE solvers (Euler, Heun, RK4) using L1 and MD metrics across the three rainfall scenarios.}
\label{tab:ode}
\vskip 0.15in
\begin{center}
\begin{small}
\begin{sc}
\begin{tabular}{ c cc cc cc }
\toprule
{Rainfall Type $\rightarrow$} & \multicolumn{2}{c}{Uniform} & \multicolumn{2}{c}{Non-uniform} & \multicolumn{2}{c}{Real-event} \\ \cmidrule(lr){2-3} \cmidrule(lr){4-5} \cmidrule(lr){6-7}
 sampler & L1 & MD & L1 & MD & L1 & MD \\ 
\midrule
Euler & 0.4 & 13.8 & 1.3 & 23.2 & 0.2 & 7.2  \\
Heun  & 0.4 & 13.6 & 1.3 & 23.1 & 0.2 & 7.0  \\
RK4   & 0.4 & 13.7 & 1.3 & 23.1 & 0.2 & 6.9  \\
\bottomrule
\end{tabular}
\end{sc}
\end{small}
\end{center}
\vskip -0.1in
\end{table*}

\paragraph{Generation Speed}

Table \ref{tab:generation_speed} presents a comparison of the average generation speeds for PIFF against two classical physics-based methods, TUFLOW and SPM. 
PIFF demonstrates significantly faster generation speeds ($\sim$ 0.1 seconds per image) compared to classical methods like TUFLOW ($\sim$ 15 seconds) and SPM ($\sim$ 3-5 seconds), offering a substantial computational advantage for real-time flood prediction. While PIFF requires an initial training phase, it's crucial to recognize that classical physics-based methods also demand considerable expert knowledge and setup for each new simulation, including defining computational grids and configuring PDEs. Thus, PIFF's rapid inference speed, following initial training, provides a distinct operational advantage.

\begin{table}[t!]
\centering
\caption{Generation Speed Comparison}
\label{tab:generation_speed}
\begin{center}
\begin{small}
\begin{sc}
\begin{tabular}{l c}
\toprule
\textbf{Method} & \textbf{Average Generation Time per Image} \\
\midrule
TUFLOW & $\sim$15 seconds \\
SPM & $\sim$3-5 seconds \\
PIFF & $\sim$0.1 seconds \\
\bottomrule
\end{tabular}
\end{sc}
\end{small}
\end{center}
\end{table}
\section{Conclusion}
\label{sec:conclusion}
We propose PIFF to enhance both efficiency and accuracy in flood depth prediction compared to traditional methods. Flow-based generative models have proven to be well-suited for high-quality flood depth map generation. Leveraging this, PIFF employs an image-to-image framework, providing more accurate predictions for both flood extent and depth. Our experiments further highlight the significant benefit of incorporating a physics-informed prior into the training process, a benefit observed through visual inspection and robust quantitative evaluations against other models across uniform, non-uniform, and real-event rainfall types. The generated flood maps facilitate effective and timely disaster mitigation, improving preparedness and response for flood management. Future work will focus on evaluating the model on different DEMs will help assess its generalizability across diverse terrains.
{
    \small
    \bibliographystyle{ieeenat_fullname}
    \bibliography{main}

@article{HallFinancialLoss,
author = {Hallegatte, Stephane and Green, Colin and Nicholls, Robert J. and Corfee-Morlot, Jan},
title = {Future flood losses in major coastal cities},
journal = {Nature Climate Change},
volume = 3,
number = 9,
pages = {802--806},
year = 2013
}

@inproceedings{Doshi2018,
  author = {Doshi, J. and Basu, S. and Pang, G.},
  title = {From Satellite Imagery to Disaster Insights},
  booktitle = {Proceedings of the 32nd Conference on Neural Information Processing Systems Workshop},
  address = {Montréal, QC, Canada},
  date = {2018-12-03/2018-12-08},
  year = {2018}
}

@Article{PengPSNET,
AUTHOR = {Peng, Bo and Meng, Zonglin and Huang, Qunying and Wang, Caixia},
TITLE = {Patch Similarity Convolutional Neural Network for Urban Flood Extent Mapping Using Bi-Temporal Satellite Multispectral Imagery},
JOURNAL = {Remote Sensing},
VOLUME = {11},
YEAR = {2019},
}

@Article{AsDCNN,
AUTHOR = {Gebrehiwot, Asmamaw and Hashemi-Beni, Leila and Thompson, Gary and Kordjamshidi, Parisa and Langan, Thomas E.},
TITLE = {Deep Convolutional Neural Network for Flood Extent Mapping Using Unmanned Aerial Vehicles Data},
JOURNAL = {Sensors},
VOLUME = {19},
YEAR = {2019},
NUMBER = {7},
URL = {https://www.mdpi.com/1424-8220/19/7/1486},
}

@article{FengRF,
  author = {Feng, Q. and Liu, J. and Gong, J.},
  title = {Urban Flood Mapping Based on Unmanned Aerial Vehicle Remote Sensing and Random Forest Classifier: A Case of Yuyao, China},
  journal = {Water},
  volume = {7},
  pages = {1437--1455},
  year = {2015},
  doi = {10.3390/w7041437}
}

@article{LIselfCNN,
title = {Urban flood mapping with an active self-learning convolutional neural network based on TerraSAR-X intensity and interferometric coherence},
journal = {ISPRS Journal of Photogrammetry and Remote Sensing},
volume = {152},
pages = {178-191},
year = {2019},
issn = {0924-2716},
doi = {https://doi.org/10.1016/j.isprsjprs.2019.04.014},
author = {Yu Li and Sandro Martinis and Marc Wieland},
}

@Article{NemniCNNSAR,
AUTHOR = {Nemni, Edoardo and Bullock, Joseph and Belabbes, Samir and Bromley, Lars},
TITLE = {Fully Convolutional Neural Network for Rapid Flood Segmentation in Synthetic Aperture Radar Imagery},
JOURNAL = {Remote Sensing},
VOLUME = {12},
YEAR = {2020},
NUMBER = {16},
}

@ARTICLE{Hash2021,
  author={Hashemi-Beni, Leila and Gebrehiwot, Asmamaw A.},
  journal={IEEE Journal of Selected Topics in Applied Earth Observations and Remote Sensing}, 
  title={Flood Extent Mapping: An Integrated Method Using Deep Learning and Region Growing Using UAV Optical Data}, 
  year={2021},
  volume={14},
  number={},
  pages={2127-2135},
  doi={10.1109/JSTARS.2021.3051873}}

@article{Ho2020DDPM,
  author = {Jonathan Ho and Ajay Jain and Pieter Abbeel},
  title = {Denoising Diffusion Probabilistic Models},
  journal = {Advances in Neural Information Processing Systems},
  volume = {33},
  pages = {6840--6851},
  year = {2020}
}

@article { VisualizingMultipleMeasuresofForecastQuality,
      author = "Paul J.  Roebber",
      title = "Visualizing Multiple Measures of Forecast Quality",
      journal = "Weather and Forecasting",
      year = "2009",
      publisher = "American Meteorological Society",
      address = "Boston MA, USA",
      volume = "24",
      number = "2",
      doi = "10.1175/2008WAF2222159.1",
      pages=      "601 - 608",
      url = "https://journals.ametsoc.org/view/journals/wefo/24/2/2008waf2222159_1.xml"
}

@article{Song2020DDIM,
  author = {Jiaming Song and Chenlin Meng and Stefano Ermon},
  title = {Denoising Diffusion Implicit Models},
  journal = {arXiv preprint arXiv:2010.02502},
  year = {2020}
}

@Article{Marc2019,
AUTHOR = {Wieland, Marc and Martinis, Sandro},
TITLE = {A Modular Processing Chain for Automated Flood Monitoring from Multi-Spectral Satellite Data},
JOURNAL = {Remote Sensing},
VOLUME = {11},
YEAR = {2019},
NUMBER = {19},
}

@Article{Bent2022,
AUTHOR = {Bentivoglio, R. and Isufi, E. and Jonkman, S. N. and Taormina, R.},
TITLE = {Deep learning methods for flood mapping: a review of existing applications and future research directions},
JOURNAL = {Hydrology and Earth System Sciences},
VOLUME = {26},
YEAR = {2022},
NUMBER = {16},
PAGES = {4345--4378},
URL = {https://hess.copernicus.org/articles/26/4345/2022/},
DOI = {10.5194/hess-26-4345-2022}
}

@article{HOSSEINY2021,
title = {A deep learning model for predicting river flood depth and extent},
journal = {Environmental Modelling \& Software},
volume = {145},
pages = {105186},
year = {2021},
issn = {1364-8152},
doi = {https://doi.org/10.1016/j.envsoft.2021.105186},
url = {https://www.sciencedirect.com/science/article/pii/S1364815221002280},
author = {Hossein Hosseiny},
}

@article{guo2021data,
  title={Data-driven flood emulation: Speeding up urban flood predictions by deep convolutional neural networks},
  author={Guo, Zifeng and Leitao, Joao P and Sim{\~o}es, Nuno E and Moosavi, Vahid},
  journal={Journal of Flood Risk Management},
  volume={14},
  number={1},
  pages={e12684},
  year={2021},
}

@article{Bradbrook01092004,
author = {K.F. Bradbrook, S.N. Lane, S.G. Waller and P.D. Bates},
title = {Two dimensional diffusion wave modelling of flood inundation using a simplified channel representation},
journal = {International Journal of River Basin Management},
volume = {2},
number = {3},
pages = {211--223},
year = {2004},
}

@inproceedings{chen2007,
author = {Chen, Albert and Djordjević, Slobodan and Leandro, J. and Savic, Dragan},
year = {2007},
month = {01},
pages = {465-472},
title = {The urban inundation model with bidirectional flow interaction between 2D overland surface and 1D sewer networks},
}

@inproceedings{lhomme2008,
author = {Lhomme, Julien and Sayers, Paul and Gouldby, Ben and Samuels, Paul and Wills, Martin},
year = {2008},
month = {10},
pages = {15-24},
title = {Recent development and application of a rapid flood spreading method},
isbn = {978-0-415-48507-4},
doi = {10.1201/9780203883020.ch2}
}

@article{LOWE2021126898,
title = {U-FLOOD – Topographic deep learning for predicting urban pluvial flood water depth},
journal = {Journal of Hydrology},
volume = {603},
pages = {126898},
year = {2021},
issn = {0022-1694},
doi = {https://doi.org/10.1016/j.jhydrol.2021.126898},
author = {Roland Löwe and Julian Böhm and David Getreuer Jensen and Jorge Leandro and Søren Højmark Rasmussen},
}

@ARTICLE{yokoya2022,
  author={Yokoya, Naoto and Yamanoi, Kazuki and He, Wei and Baier, Gerald and Adriano, Bruno and Miura, Hiroyuki and Oishi, Satoru},
  journal={IEEE Transactions on Geoscience and Remote Sensing}, 
  title={Breaking Limits of Remote Sensing by Deep Learning From Simulated Data for Flood and Debris-Flow Mapping}, 
  year={2022},
  volume={60},
  number={},
  pages={1-15}}

@ARTICLE{Lütjens2024,
  author={Lütjens, Björn and Leshchinskiy, Brandon and Boulais, Océane and Chishtie, Farrukh and Díaz-Rodríguez, Natalia and Masson-Forsythe, Margaux and Mata-Payerro, Ana and Requena-Mesa, Christian and Sankaranarayanan, Aruna and Piña, Aaron and Gal, Yarin and Raïssi, Chedy and Lavin, Alexander and Newman, Dava},
  journal={IEEE Transactions on Geoscience and Remote Sensing}, 
  title={Generating Physically-Consistent Satellite Imagery for Climate Visualizations}, 
  year={2024},
  volume={62},
  number={},
  pages={1-11},
  doi={10.1109/TGRS.2024.3493763}}

@article{Jones2023,
  author    = {Jones, A. and Kuehnert, J. and Fraccaro, P. and others},
  title     = {AI for climate impacts: applications in flood risk},
  journal   = {npj Climate and Atmospheric Science},
  volume    = {6},
  pages     = {63},
  year      = {2023},
  doi       = {10.1038/s41612-023-00388-1}
}

@article{seo2023improved,
  title={Improved flood insights: Diffusion-based sar to eo image translation},
  author={Seo, Minseok and Oh, Youngtack and Kim, Doyi and Kang, Dongmin and Choi, Yeji},
  journal={arXiv preprint arXiv:2307.07123},
  year={2023}
}

@Article{floodgan2021,
AUTHOR = {Hofmann, Julian and Schüttrumpf, Holger},
TITLE = {floodGAN: Using Deep Adversarial Learning to Predict Pluvial Flooding in Real Time},
JOURNAL = {Water},
VOLUME = {13},
YEAR = {2021},
NUMBER = {16},
ARTICLE-NUMBER = {2255},
}

@Article{yang2015spm,
AUTHOR = {Yang, Tsun-Hua and Chen, Yi-Chin and Chang, Ya-Chi and Yang, Sheng-Chi and Ho, Jui-Yi},
TITLE = {Comparison of Different Grid Cell Ordering Approaches in a Simplified Inundation Model},
JOURNAL = {Water},
VOLUME = {7},
YEAR = {2015},
NUMBER = {2},
PAGES = {438--454},
URL = {https://www.mdpi.com/2073-4441/7/2/438},
ISSN = {2073-4441},
DOI = {10.3390/w7020438}
}

@article{heusel2017gans,
  title={Gans trained by a two time-scale update rule converge to a local nash equilibrium},
  author={Heusel, Martin and Ramsauer, Hubert and Unterthiner, Thomas and Nessler, Bernhard and Hochreiter, Sepp},
  journal={Advances in neural information processing systems},
  volume={30},
  year={2017}
}

@misc{ronneberger2015unet,
      title={U-Net: Convolutional Networks for Biomedical Image Segmentation}, 
      author={Olaf Ronneberger and Philipp Fischer and Thomas Brox},
      year={2015},
      eprint={1505.04597},
      archivePrefix={arXiv},
      primaryClass={cs.CV},
      url={https://arxiv.org/abs/1505.04597}, 
}

@misc{isola2018gan,
      title={Image-to-Image Translation with Conditional Adversarial Networks}, 
      author={Phillip Isola and Jun-Yan Zhu and Tinghui Zhou and Alexei A. Efros},
      year={2018},
      eprint={1611.07004},
      archivePrefix={arXiv},
      primaryClass={cs.CV},
      url={https://arxiv.org/abs/1611.07004}, 
}

@article{lipman2022flow,
  title={Flow matching for generative modeling},
  author={Lipman, Yaron and Chen, Ricky TQ and Ben-Hamu, Heli and Nickel, Maximilian and Le, Matt},
  journal={arXiv preprint arXiv:2210.02747},
  year={2022}
}

@article{tong2023improving,
  title={Improving and generalizing flow-based generative models with minibatch optimal transport},
  author={Tong, Alexander and Fatras, Kilian and Malkin, Nikolay and Huguet, Guillaume and Zhang, Yanlei and Rector-Brooks, Jarrid and Wolf, Guy and Bengio, Yoshua},
  journal={arXiv preprint arXiv:2302.00482},
  year={2023}
}

@article{vaswani2017attention,
  title={Attention is all you need},
  author={Vaswani, Ashish and Shazeer, Noam and Parmar, Niki and Uszkoreit, Jakob and Jones, Llion and Gomez, Aidan N and Kaiser, {\L}ukasz and Polosukhin, Illia},
  journal={Advances in neural information processing systems},
  volume={30},
  year={2017}
}

@article{li2022satellite,
  title={Satellite detection of surface water extent: A review of methodology},
  author={Li, Jiaxin and Ma, Ronghua and Cao, Zhigang and Xue, Kun and Xiong, Junfeng and Hu, Minqi and Feng, Xuejiao},
  journal={Water},
  volume={14},
  number={7},
  pages={1148},
  year={2022},
  publisher={MDPI}
}

@article{uddin2019operational,
  title={Operational flood mapping using multi-temporal Sentinel-1 SAR images: A case study from Bangladesh},
  author={Uddin, Kabir and Matin, Mir A and Meyer, Franz J},
  journal={Remote Sensing},
  volume={11},
  number={13},
  pages={1581},
  year={2019},
  publisher={MDPI}
}

@article{lin2023rapid,
  title={Rapid urban flood risk mapping for data-scarce environments using social sensing and region-stable deep neural network},
  author={Lin, Lin and Tang, Chaoqing and Liang, Qiuhua and Wu, Zening and Wang, Xinling and Zhao, Shan},
  journal={Journal of Hydrology},
  volume={617},
  pages={128758},
  year={2023},
  publisher={Elsevier}
}

@article{ming2020real,
  title={Real-time flood forecasting based on a high-performance 2-D hydrodynamic model and numerical weather predictions},
  author={Ming, Xiaodong and Liang, Qiuhua and Xia, Xilin and Li, Dingmin and Fowler, Hayley J},
  journal={Water Resources Research},
  volume={56},
  number={7},
  pages={e2019WR025583},
  year={2020},
  publisher={Wiley Online Library}
}

@misc{tuflow,
  author       = {BMT},
  title        = {TUFLOW},
  year         = 2025,
  url          = {https://www.tuflow.com/},
  note         = {Version 2025-01}
}

@article{chiang2024efficient,
  title={An efficient 2-D flood inundation modelling based on a data-driven approach},
  author={Chiang, Shen and Fu, Huei-Shuin and Hsiao, Shih-Chun and Hsiao, Yi-Hua and Chen, Wei-Bo},
  journal={Journal of Hydrology: Regional Studies},
  volume={52},
  pages={101741},
  year={2024},
  publisher={Elsevier}
}

@article{wijaya2023rapid,
  title={A rapid flood inundation model for urban flood analyses},
  author={Wijaya, Obaja Triputera and Yang, Tsun-Hua and Hsu, Hao-Ming and Gourbesville, Philippe},
  journal={MethodsX},
  volume={10},
  pages={102202},
  year={2023},
  publisher={Elsevier}
}

@article{kumar2025machine,
  title={Machine learning applications in flood forecasting and predictions, challenges, and way-out in the perspective of changing environment},
  author={Kumar, Vijendra and Sharma, Kul Vaibhav and Mangukiya, Nikunj K and Tiwari, Deepak Kumar and Ramkar, Preeti Vijay and Rathnayake, Upaka},
  journal={AIMS Environmental Science},
  volume={12},
  number={1},
  pages={72--105},
  year={2025}
}
}


\end{document}